# A generalized decision tree ensemble based on the NeuralNetworks architecture: Distributed Gradient Boosting Forest (DGBF)

Ángel Delgado-Panadero[1] · José Alberto Benítez-Andrades[2]  · María Teresa García-Ordás[3]

## Abstract
Tree ensemble algorithms as RandomForest and GradientBoosting are currently the dominant methods for modeling discrete or tabular data, however, they are unable to perform a hierarchical representation learning from raw data as NeuralNetworks does thanks to its multi-layered structure, which is a key feature for DeepLearning problems and modeling unstructured data. This limitation is due to the fact that tree algorithms can not be trained with back-propagation because of their mathematical nature. However, in this work, we demonstrate that the mathematical formulation of bagging and boosting can be combined together to define a graph-structured-tree-ensemble algorithm with a distributed representation learning process between trees naturally (without using back-propagation). We call this novel approach Distributed Gradient Boosting Forest (DGBF) and we demonstrate that both RandomForest and GradientBoosting can be expressed as particular graph architectures of DGBT. Finally, we see that the distributed learning outperforms both RandomForest and GradientBoosting in 7 out of 9 datasets.

## 1 Introduction

Tree ensemble methods have become one of the most successful and widely-used algorithms in machine learning to model discrete or tabular data in a variety in both academia and industry [1, 2]. One of the reasons is that they are built on CART model [3], which is a global approximator function (i.e. it is able to approximate any non-stochastic function in the infinite dataset size limit). However, one of the handicaps of tree ensemble models is that they are not able to learn a good representation from raw data as Neural Networks (NN) does [4].

The deep-graph or hierarchical structure of NN does not only allow representational learning from raw data, but it also seems to be one of the key ingredients of their success as a global approximator [5]. Because of that, many works have attempted to give the tree ensemble algorithms NN properties such as a multi-layered structure or a distributed representation learning [6, 7]. However, in the majority of the cases, they are not able to introduce these properties naturally in the tree ensemble because of the non-derivable and non-parametric nature of CARTs. Instead, they try to imitate the back-propagation algorithm by modifying the mathematical formulation of CART or the tree ensemble algorithm to make it parametric.

In real-world problems, with finite data and stochastic contribution, CARTs have shown to be biased because of the high variance of their predictions. Tree ensemble models, such as RandomForest [8] and GradientBoosting [9] are able to solve the CART bias by gathering the predictions of many trees together. Despite of the improvement and their great results, tree ensembles have also shown to produce biased predictions depending on their implementation [10, 11]. For the

José Alberto Benítez-Andrades and María Teresa García-Ordás contributed equally to this work.

✉ José Alberto Benítez-Andrades
  jbena@unileon.es

  Ángel Delgado-Panadero
  delgadopanadero@gmail.com

  María Teresa García-Ordás
  mgaro@unileon.es

[1] Paradigma Digital S.L., Vía de las dos Castillas, 33, Pozuelo de Alarcón 28224, Madrid, Spain

[2] SALBIS Research Group, Department of Electric, Systems and Automatics Engineering, Universidad de León, Campus of Vegazana s/n, León 24071, León, Spain

[3] SECOMUCI Research Group, Escuela de Ingenierías Industrial e Informática, Universidad de León, Campus of Vegazana s/n, León 24071, León, Spain

latter, many solutions have been proposed based on combining *boosting* and *bagging* algorithms together. The majority of them have shown an improvement from using *bagging* and *boosting* separately, but they have not made a mathematical interpretation of the reasons for this improvement in terms of statistical learning.

In this paper, we are going to analyze the mathematical formulation of RandomForest and GradientBoosting to propose a more general tree ensemble algorithm that combines *boosting* and *bagging* together, Distributed Gradient Boosting Forest (DGBF), leading to a distributed graph-structured model analog to a Dense Neural Network. We are going to demonstrate that RandomForest and GradientBoosting can be understood as particular cases of DGBF. Finally, we will test the results of DGBF over nine experiments with nine regression datasets against the results from RandomForest and GradientBoosting.

## 2 Related work

Tree ensemble algorithms are based mainly on two techniques: *bagging* and *boosting*. RandomForest is an ensemble algorithm based on *bagging* where each tree of the ensemble is trained on a bootstrap subsample of the original dataset which leads to a bias reduction in the predictions of the ensemble from the predictions of the individual learners. In contrast to other algorithms [12] have demonstrated that the predictions from RandomForest follow a U-statistic, which means that the error deviation is limited to a known and finite confident interval. RandomForest outperforms most state-of-the-art learners on high dimensional data [2] and their structure allows a learning parallelization process. The downside is that even though RandomForest has demonstrated to reduce the bias on the predictions, it lacks an optimization learning process between trees which leads to biased predictions [11].

In contrast, *boosting* models perform an optimization process through an additive inclusion of learners to the ensemble where each learner performs a gradient descent of the previous predictions of the ensemble. GradientBoosting builds a hierarchical sequence of trees where each one is fitted with the gradient descent of the predictions of the previous trees. Despite of the hierarchical learning process of GradientBoosting, it also lacks a mechanism to efficiently learn internal representations as NN [5]. In contrast with RF, this approach does not ensure the variance reduction and tends to overfit and be biased because of the finite dataset size of the real-world scenario [10]. In the majority of the cases, this overfitting is overcome by reducing the learning capacity of each *boosting* step by adding a learning rate weight or by reducing the learning capacity of the weak learner, for instance, reducing the max depth with CARTs [13].

Despite being different approaches, *boosting* and *bagging* are not incompatible, in fact, they can be used jointly, however, there is not a unique solution for how to combine them. In most cases, it is based on making *bagging* of *boosting* ensembles or *boosting* of *bagging* ensembles [14, 15]. Even though they give good results, in the majority of the cases there is not a mathematical interpretation of the learning representation. In [16] they make an analysis of the prediction distribution in RandomForest to analyze the stability of the learning process applying boosting and the variance of the predictions. These approaches show an improvement of the ensemble by combining *boosting* and bagging, but they do not offer distributed learning or representational learning.

Regarding NN, many people think that their representative learning is based on their deep hierarchical structure and past works have tried to dotate tree ensemble algorithms of this feature [17, 18]. For doing so, the main ideas pass through applying the back-propagation algorithm with trees, for instance by transforming the tree function structure to make it derivable [7]. In [6] they also try to combine it with the routing structure of the tree predictions. Other works, try to build a tree inverse functions, instead of modifying the tree structure [19]. These approaches are able to make representational learning for the tree ensemble through modifying the tree or *boosting* algorithm, but none of them is able to make it naturally.

In the last few year RF and GBDT have shown to be two of the most powerful algorithms in Machine Learning for tabular data. It is widely used in many fields and its rate of success is very high: healthcare [20–22], education [23], energy [24, 25], economics [26, 27], etc. Recently it has also been proven that tree ensemble algorithm are locally explainable [28] as well as global explainable [29].

Our motivation is not based on imitating NN algorithm or combining techniques of *bagging* and *boosting* experimentally but rather, we will analyze mathematically the combination of *boosting* and bagging. We will see that for the RMSE loss function, this combination leads to a distributed representation of the gradient descent in the contribution of the gradient for each tree. Under this mathematical formulation, we can define a tree ensemble with a deep graph structure, where each tree of the ensemble learns a concrete representation of the gradient descent.

### 2.1 Background in tree ensembles

Ensembles are a kind of machine learning model that join the prediction from a set of weak learners to produce predictions statistically more powerful than those from the weak learners alone. RandomForest and GradientBoosting are two ensemble models which generally use CART as a weak learner. CART function is given by

$$h(x; \{b_j, R_j\}_1^J) = \sum_{j=0}^{J} b_j 1(x \in R_j)$$
$$= \sum_{j=0}^{J} E[y_i \mid x_i]_{x_i \in R_j} 1(x \in R_j). \quad (1)$$

This is the CART formula [3], where the parameter $J$ is the number of nodes of the tree $R_j$ is the terminal node (the index $j$ does not reflect any particular order). The parameter $b_j$ is the node value. The expression (1) is very powerful and can approximate any function, however, because of the finite-data problem in the training process, it produces two kinds of biases (see Appendix ). One way of combining the prediction of many trees to reduce bias is to aggregate the results by the mean. Given a dataset, the learning process of a tree is deterministic, so to make multiple trees produce different predictions, each of them is trained in a different bootstrap subsample of the dataset. This technique is called *bagging*.

$$RF(x) = \frac{1}{T} \sum_{t=0}^{T} h_t(x), \quad (2)$$

where $T$ is the number of trees of the ensemble, and $h_t(x)$ is trained to minimize the loss function, $L(y, f(x))$, over a *bagging* subsample $\{x\}_t$

$$h_t(x) = \arg\min_{h} L(y_i, h(x_i))/x_i \in \{x\}_t. \quad (3)$$

This technique is based on the central limit theorem from statistics where we expect that the variance manifest of the CART biases can be reduced by averaging over enough tree predictions. In contrast, with GradientBoosting each tree does not try to learn over a different dataset subsample, but it follows a "stage-wise" optimization approach, where each tree is added to the ensemble to reduce the global loss from the previous trees. The final prediction is the sum of all the trees in the ensemble. This process is called *boosting* [9]

$$F_l(x) = \sum_{l=0}^{L} h_l(x), \quad (4)$$

where $L$ is the number of trees of the ensemble. To reduce the loss function from the previous trees, $L(y, F_{l-1}(x))$, each of those trees is fitted with the gradient of the loss function from the previous predictors In real-world problems, we never have the density function, consequently, the *boosting* is approximated using finite data by assuming regularity in the $P(y \mid x)$ distribution. To do so, each tree is trained with pseudo-responses that are computed (in the case of the $RMSE$ loss) as the difference between the target and the predictions of a current ensemble of trees (residual errors).

$$g_l(x) = E_y \left[ \frac{\partial L(y, F(x))}{\partial F(x)} \right]_{F=F_{l-1}} \equiv y - F_{l-1}(x), \quad (5)$$

$$h_l(x) \simeq \arg\min_{h'} E\left[ L(g_{l-1}(x), h'(x)) \right], \quad (6)$$

## 3 Distributed gradient boosting forest: DGBF

### 3.1 Mathematical formulation

Noting the lack of a real optimization process in the Random-Forest algorithm, our attempt is going to define a gradient descent ensemble using GradientBoosting and RandomForest as a weak learner. This function can be mathematically expressed as:

$$F_l(x) = \sum_{l=1}^{L} RF_l(x) = \frac{1}{T} \sum_{l=0}^{L} \sum_{t=0}^{T} h_{l,t}(x) \quad F_0(x) = E[y_i], \quad (7)$$

where $L$ is the number of *boosting* steps and $T$ the number of trees for each RandomForest.

#### 3.1.1 Splitting gradients

Rather than using the residuals from the whole ensemble, our goal is to apply (5) over each element of the RandomForest in each *boosting* step with the idea to distribute the learning optimization process between all the weak learners. Following the mathematical definition of *boosting* (5), when the weak learner can be expressed as the sum of functions (for instance a RandomForest), the gradient can be expressed as the sum of the gradients of each function.

$$g_l(x_i) = y_i - F_{l-1}(x_i) = y_i - \sum_{l'=0}^{l-1} RF_{l'}(x_i)$$
$$= \sum_{l'=0}^{l-1} \sum_{t=0}^{T} \frac{1}{T}(T * y_i - h_{l',t}(x_i)) \equiv \frac{1}{T} \sum_{t=0}^{T} g'_{l,t}(x_i). \quad (8)$$

The previous equation does not only show a mathematical identity, but it also gives a new optimization process where we do not try to make a gradient descent of the whole RandomForest gradients, $g_l(x_i)$, but rather, make a gradient descent of each tree, $g'_{l,t}(x_i)$, independently.

### 3.1.2 Distributed learning

The next *boosting* step trains a new RandomForest model using the gradients components $g'_{l,t}(x_i)$. Taking advantage of the multi-target learning capability of CARTs, each tree of the RandomForest ensemble can learn simultaneously all the gradient components independently.

$$h_{l,t}(x) = \arg\min_h \sum_{t'=0}^{T} L(g'_{l-1,t'}(x_i), h(x_i))/x_i \in \{x\}_{l,t}. \quad (9)$$

Once the tree is trained to model $g'_{l,t}(x_i)$ from the formula (8) we can conclude that the estimation for the actual gradient $g'_l(x_i)$ can be estimated as the mean value of all the outputs of the tree. However, in our implementation, to avoid the prediction being biased by the subsample used for training, we compute a weighted average of the tree outputs, where the weights are computed to minimize the global loss function.

$$h_{l,t}(x) = \sum_{t'=0}^{T} w_{t'} \cdot h_{l,t}^{t'}(x). \quad (10)$$

$$w_t = \arg\min_w L\left(g_{l-1}(x_i), \sum_{t'=0}^{T} w_{t'} \cdot h_{l,t}^{t'}(x)\right), \quad (11)$$

where $h_{l,t}^{t'}(x)$ is the output $t'$ of the tree $h_{l,t}(x)$. Under this formulation, the tree ensemble follows a distributed analog to a Dense Neural Network model, as can be seen in Fig. 1, consequently, from now on, we will refer to each RandomForest from each step as a "tree layer".

Even though distributed learning has been studied before, the contrubution of our novel approach is not to make a distributed learning of the target directly, but rather the distribution of each gradient of the previous trees independently by using the splitting gradient technique from Section 3.1.1.

### 3.1.3 Dynamic sampling

As a conventional RandomForest, each tree is trained with a different bootstrap subsample of the dataset, these subsamples are neither the same from one *boosting* step to another. From the ideas of *dynamic boosting* from [10], we should begin training each of the trees from the beginning layers with a small subsample, and increase it from one *boosting* to another.

$$h_{l,t}(x) = \arg\min_h L(y_i - B_{l-1}, h(x_i))/x_i \in \{x\}_{l,t} \quad (12)$$

$$\{x\}_{l=0,t} < \{x\}_{l=l',t} < \{x\}_{l=L,t},$$

where, $\{x\}_{l,t}$ is the dataset sample for the tree $t$ in the layer $l$. From these ideas, we could try to generate the datasets from the trees of the same layer through *bagging*, however, in our case we have used a split of the original dataset in equally-sized subsamples to train each of the trees. We preferred this approach rather than *bagging* because we ensure that all the data samples from the original dataset are used and have the same weight during training. From one layer to another the size of the sample of each tree is increased with new data, until the last layer, which will use a sample with the majority of the data for each tree.

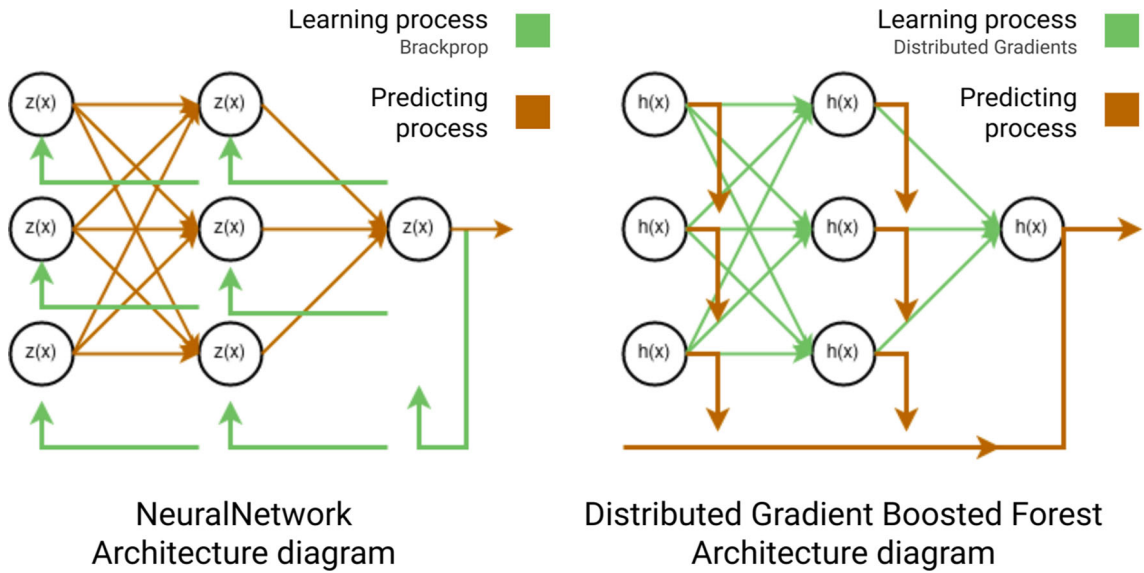

**Fig. 1** Diagram with NN and DGBF predicting and learning algorithm. In NN, the output of each neuron is used as the input of the next layer neuron, while in the DGBF, the prediction of the whole trees is added and averaged together. In the learning process, NN, update iteratively the weights of the neurons with the back-propagation algorithm, while DGBF, forwards, the distributed gradients of all the tree from each layer, to every tree of the following layer

**Fig. 2** Representantion of DGBF architecture in the particular cases where we recover the RandomForest and GradientBoosting algorithms

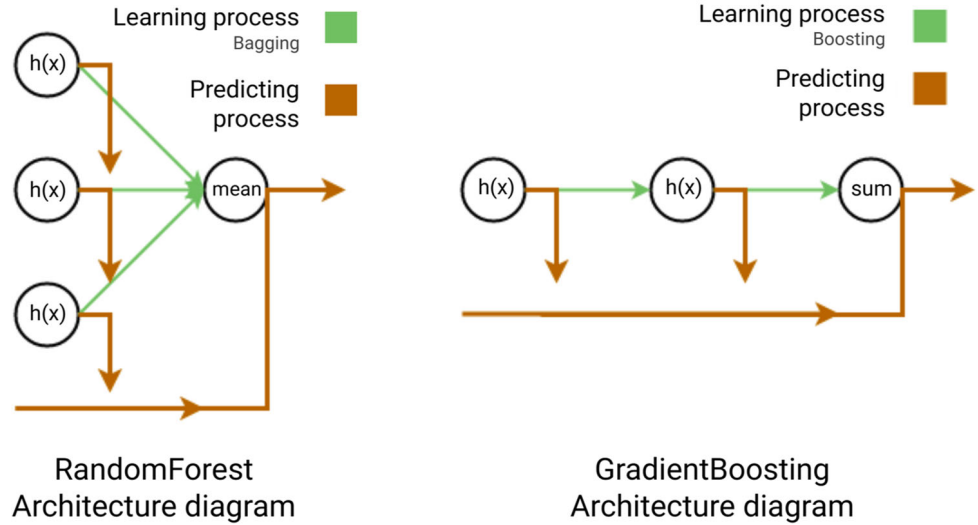

RandomForest Architecture diagram

GradientBoosting Architecture diagram

Dynamic sampling is commonly used to avoid overfitting during the boosting process, however in the context of the splitting gradients approach, its aim is to achieve more possible splitting regions, which helps the algorithm to increase the learning capacity. The problem of spliting regions is explained with more detail in the Appendix .

### 3.2 Derivation of randomforest and gradientboosting from DGBF formula

It is important to note that the optimization formula proposed in (9) is not a substitution of the conventional *boosting* formula defined in (6), but rather, it is a more general case where the weak estimator can be expressed as a sum of functions. In the particular case when the sum has only one term, we recover the conventional GradientBoosting.

$$F_l(x) = \frac{1}{T}\sum_{l=0}^{L}\sum_{t=0}^{T}h_{l,t}(x) \xrightarrow{T=1} F_l(x) = \sum_{l=0}^{L}h_l(x), \quad (13)$$

$$\arg\min_h \sum_{t'=0}^{T} L(g'_{l-1,t'}(x_i), h(x_i)) \xrightarrow{T=1}$$
$$\arg\min_h L(r_{l-1}(x_i), h(x_i)). \quad (14)$$

Moreover, the RandomForest algorithm can be also recovered, in the particular case when we only have one layer

$$F_l(x) = \sum_{l=1}^{L} RF_l(x) \xrightarrow{L=1} F(x) = RF(x). \quad (15)$$

For simplicity we have written the $F_l(x)$ without the prior estimator $F_0$, however, it does not affect the result because the RandomForest learning is invariant by and addition of a constant value in the target. In Fig. 2 we can see how

**Algorithm 1** Training process for DGBF.

**Require:** $L, T, lr, X, y$
$F_0 \leftarrow mean(y)$
$F_{l,t} \leftarrow new\ DecisionTree$
**for** $l = 0 : L$ **do**  ▷ Loop for creating each layer of the ensemble
  $pred_{l',t'} \leftarrow Initialize()$
  **for** $l' = 0 : l$ **do**  ▷ Loop for the previous layers
    **for** $t' = 0 : T$ **do**  ▷ Loop for each tree of the layer
      $pred_{l',t'} \leftarrow pred_{l',t'} + \sum_{r=0}^{T} w^r_{l',t'} * F^r_{l',t'}(X);$
    **end for**
  **end for**
  $g_{l',t'} \leftarrow y - pred_{l',t'}$  ▷ Compute residuals of each tree
  $g_{l',t'} \leftarrow lr * g_{l',t'}$  ▷ Learning rate
  **for** $t' = 0 : T$ **do**  ▷ Loop over current layer trees
    $x^i_{l',t'}\ g^i_{l',t'} \leftarrow BootstrapSample(X,\ g_{l',t'});$
    $F_{l,t} \leftarrow FitTree(F_{l,t}, x^i_{l',t'}, g^i_{l',t'});$  ▷ Train with the splitted gradient
    $w^r_{l',t'} \leftarrow ComputeWeights(F_{l,t}, X, g_{l',t'});$  ▷ Weights for prediction outputs.
  **end for**
**end for**

RandomgForest and GradientBoosting can be represented as special cases the diagram for DGBF in Fig. 1.

## 4 Experiments and results

### 4.1 Experiments conditions

In these experiments we are going to compare the results of our model with the results from GradientBoosting and RandomForest, trained over the 9 datasets from Table 1. To make the comparison results equivalent when comparing the three models, we are going to use the same number of estimators for all of them (100). The GradientBoosting and RandomForest implementation that we are going to use is from the library Scikit-Learn with with all the other parameters are set as default in the Scikit-Learn v1.0.2 implementatio). For DGBF we are going to use 5 layers with 20 estimators each with a max depth in the trees of 10. To guarantee the reproducibility of the experiments carried out, the complete code is available in an open repository.[1]

For each of the datasets, we are going to split it into a train and test set (with a percentage of 80% and 20% respectively) and use the train set to train the three models. To make the metric measurement statistically significant, we are not going to measure the score over one experiment with just a train-test split for each dataset, but rather, we are going to make 200 simulations for each dataset with different randomized splits.

To interpret the results we are going to perform three kinds of statistics. One is where we compare the score distribution for the 200 simulations for both models a second one is where we plot the paired difference distribution of the score from the three models for each experiment and finally, we are going to make a test to meassure the time complexity of the training process. With the first, we are going to be able to analyze the variance and distribution of the scores of each of the models separately meanwhile the second analysis is more statistically powerful to compare both the scores of both models (RandomForest and GradientBoosting) with DGBF.

## 4.2 Experiment 1 - precision results

To compare the performance of the three models in each of the experiments, we have computed the adjusted $R$ score of the predictions of the model over the test set with its target. From the execution of the experiments, the average score over the simulations experiments with RandomForest (RF) GradientBoosting (GBDT) and our model, Distributed Gradient Boosting Forest (DGBF) can be seen in Table 2. In Fig. 3 it can be seen the distribution of scores of all the experiments.

From the results, we see that the mean score from DGBF surpasses the mean score of GradientBoosting and RandomForest in 7 out of 9 datasets and in 5 of them with a paired difference mean is greater than its standard deviation in both RandomForest and GradientBoosting, in favor to GBDT. With the number of experiments performed this means that DGBF with an architecture of 20 layers and 5 trees per layer performs better than RandomForest and GradientBoosting with high statistical significance.

For the two datasets where DGBF performs worse than RandomForest, it is important to note as we explained in Section 3.2, RandomForest and GradientBoosting are par-ticular architectures of DGBF. Consequently with recover the results of RandomForest with a DGBF architecture of just one layer. The previous results are specific for a DGBF of 20 layers and 5 trees per layer.

## 4.3 Experiment 2 - ablation test

In these experiment we are going replicate the previous experiment of 200 simulations with 9 different datasets but for each of these simulations we are going to artificialy remove two features from training set. The aim of this test is to test the robustness to overfitting when predictive variables are removed. The results can be seen in Table 3. In Fig. 4 it can be seen the distribution of scores of all the experiments.

In these experiments, instead of focusing on the precision of the paired difference, we have measured the variance of the error, to check for overfitting. We can see that in general, DGBF not only performs better in the ablation experiment than GBDT and RF, but it also shows less variance in its precision. The latter demonstrates that DGBF is stable in the ablation scenario and robust against overfitting.

## 4.4 Experiment 3 - complexity test

In this experiment, we are going to analyze the model complexity of our novel proposal and compare it with the model complexity of RF and GBDT. To do so, the aim of this experiment is to measure the execution time of each algorithm with a different size of training data. The training data is generated randomly with a fixed number of columns (specifically 5).

For GBDT and RF we use the Scikit-Learn implementation written in Cython. In the case of DGBF our implementation is written in Python. We can see in Fig. 5 that DGBF is slightly slower than RF and a more effetien implementation could reach a execution time similar to GBDT.

# 5 Future steps

The main motivation of this work is not to find an alternative ensemble algorithm that outperforms RandomForest or GradientBoosting. There are already many contributions in the literature that have demonstrated before the improvement of tree ensembles by combining together *boosting* and *bagging* techniques. The novelty of our contribution is the definition of a tree ensemble algorithm that naturally performs a distributed representation learning without the modification of the tree algorithm to be compatible with back-propagation algorithm.

---

[1] https://doi.org/10.5281/zenodo.7236216

**Table 1** Experiment datasets

| Name | Type | Sample | Area | feature types | attributes |
|---|---|---|---|---|---|
| Parkinson | Multivariate | 5875 | Life | Real | 23 |
| Wine | Multivariate | 1599 | Chemical | Real Categorical | 13 |
| Concrete | Multivariate | 1030 | Physical | Real | 8 |
| Obesity | Multivariate | 2112 | Life | Real Categorical | 17 |
| NavalVessel | Multivariate | 11934 | Computer | Real | 18 |
| Temperature | Multivariate | 7589 | Physical | Real Date | 23 |
| Cargo2000 | Multivariate | 3943 | Business | Real Int | 98 |
| BikeSales | Multivariate | 8761 | Business | Real Int Categorical | 16 |
| Superconduct | Multivariate | 21264 | Physical | Real Int | 79 |

All these datasets hava been taken from UC Irvine Machine Learning Repository. All of them have been preprocessed by removing id column, converting string categorial columns into numeric, filling NA with 0 and spliting date columns into year, month and day

**Table 2** Average scores from the 200 experiments performed for each dataset

| Data | GBDT | RF | DGBF | GBDT PairDiff | RF PairDiff |
|---|---|---|---|---|---|
| Parkinson | 0.840 | 0.848 | **0.852** | 0.011 ± 0.004 | 0.004 ± 0.004 |
| Wine | 0.40 | **0.47** | 0.45 | 0.05 ± 0.03 | -0.018 ± 0.02 |
| Concrete | 0.90 | 0.90 | **0.93** | 0.031 ± 0.008 | 0.026 ± 0.008 |
| Obesity | 0.983 | 0.989 | **0.990** | 0.006 ± 0.001 | 0.0005 ± 0.001 |
| NavalVessel | 0.857 | 0.9937 | **0.9960** | 0.138 ± 0.005 | 0.0023 ± 0.0006 |
| Temperature | 0.864 | 0.909 | **0.931** | 0.0668 ± 0.004 | 0.022 ± 0.003 |
| Cargo2000 | 0.99996 | 0.99997 | **0.99998** | 1e-05 ± 6e-5 | 1e-07 ± 3e-5 |
| BikeSales | 0.848 | 0.882 | **0.892** | 0.043 ± 0.005 | 0.009 ± 0.003 |
| Superconduc | 0.8627 | **0.9226** | 0.9221 | 0.059 ± 0.003 | -0.0005 ± 0.002 |

The model that gives the best results for each dataset has been highlighted in bold

**Table 3** Average scores from the 50 simulations performed for each dataset in the ablation scenario

| Data | GBDT | RF | DGBF |
|---|---|---|---|
| Parkinson | 0.834 ± 0.009 | 0.837 ± 0.009 | **0.841 ± 0.008** |
| Wine | 0.375 ± 0.033 | **0.443 ± 0.038** | 0.44 ± 0.043 |
| Concrete | 0.885 ± 0.011 | 0.89 ± 0.015 | **0.93 ± 0.012** |
| Obesity | 0.983 ± 0.002 | 0.989 ± 0.003 | **0.990 ± 0.002** |
| NavalVessel | 0.794 ± 0.006 | 0.9937 ± 0.0012 | **0.9960 ± 0.0007** |
| Temperature | 0.864 ± 0.007 | 0.909 ± 0.004 | **0.931 ± 0.003** |
| Cargo2000 | 0.99996 ± 4e-5 | 0.99997 ± 2e-5 | **0.99998 ± 2e-5** |
| BikeSales | 0.674 ± 0.010 | 0.748 ± 0.012 | **0.775 ± 0.010** |
| Superconduc | 0.861 ± 0.004 | **0.923 ± 0.003** | 0.922 ± 0.003 |

The model that gives the best results for each dataset has been highlighted in bold

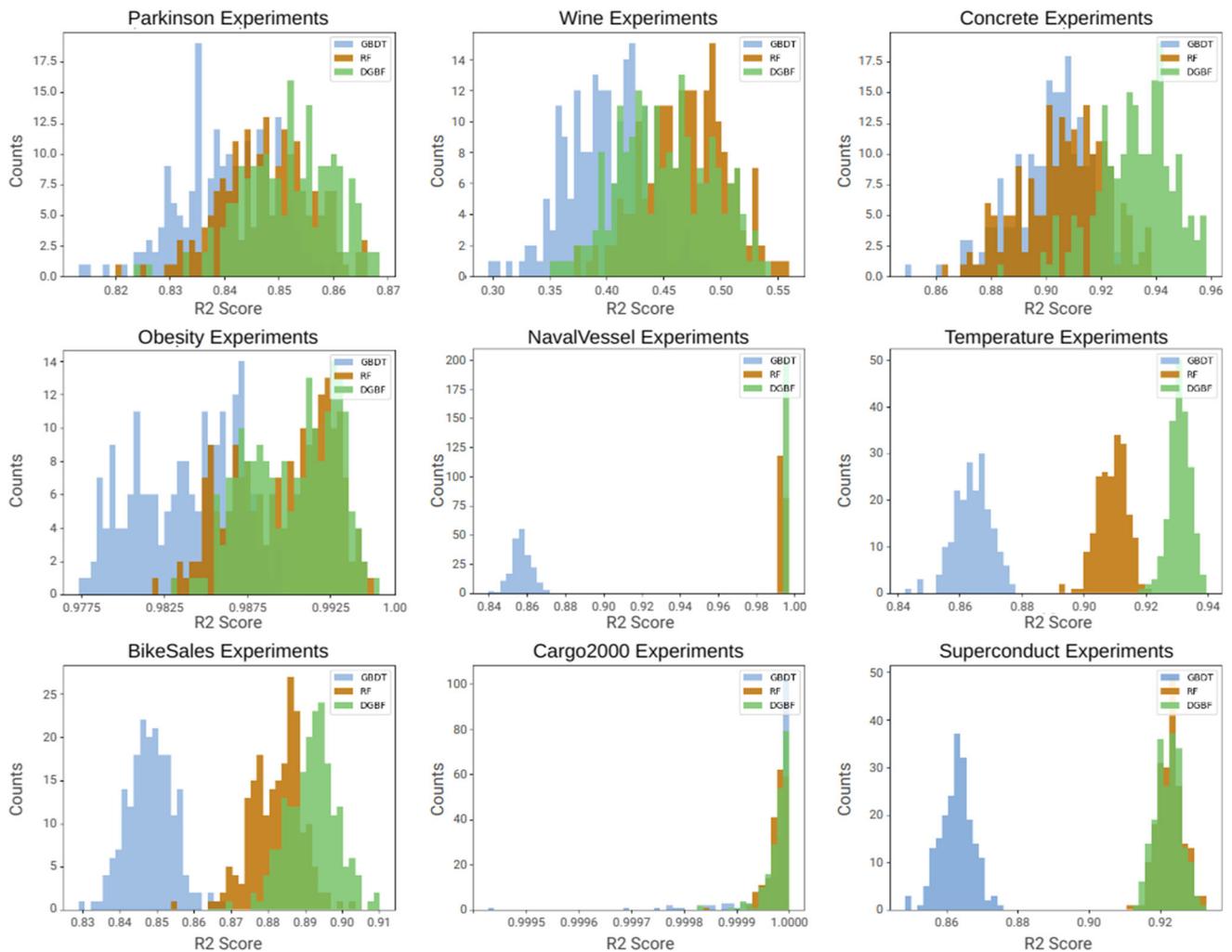

**Fig. 3** In this image it can be seen the distribution of R2 score of both models, RandomForest (brown), GradientBoosting (blue) and DGBF (green), for the 200 experiments of each dataset

This approach defines a deep-graph hierarchical ensemble analogue to a Dense Neural Network. The ability of defining tree ensembles based on a graph architecture, does not only lead to a increase in the learning capability, but it also open new researching areas applying NN strategies to tree ensemble algorithms (such as Dropout or Residual connections).

Moreover, one of the limitations of tree ensemble algorithms is their incapacity to model other data structures beyond tabular data (images, text,…). Deep graph structure have demonstrated to be the key component of NN to model such kinds of data. Our novelty approach allows the possibility to model algorithms such as text embeddings or image convolutional models.

## 6 Conclusions

In this paper, we present a novel tree ensemble algorithm: Distributed Gradient Boosting Forest that combines *boosting* and *bagging*. Its difference from other tree ensemble algorithms is that we have defined mathematically a distributed optimization process that naturally leads to a deep-hierarchical structure, analog to a Dense NeuralNetworks, where each tree learns an independent representation from raw data without NN techniques such as back-propagation or parametric models.

We have demonstrated that DGBF is not only a more general tree ensemble algorithm than RandomForest and GradientBoosting but also these two algorithms can be under-stood as particular cases of DGBF. Finally, we have seen in a set of 200 simulations with four experiments with regression datasets that DGBF equals or surpasses the learning capacity from RandomForest and GradientBoosting.

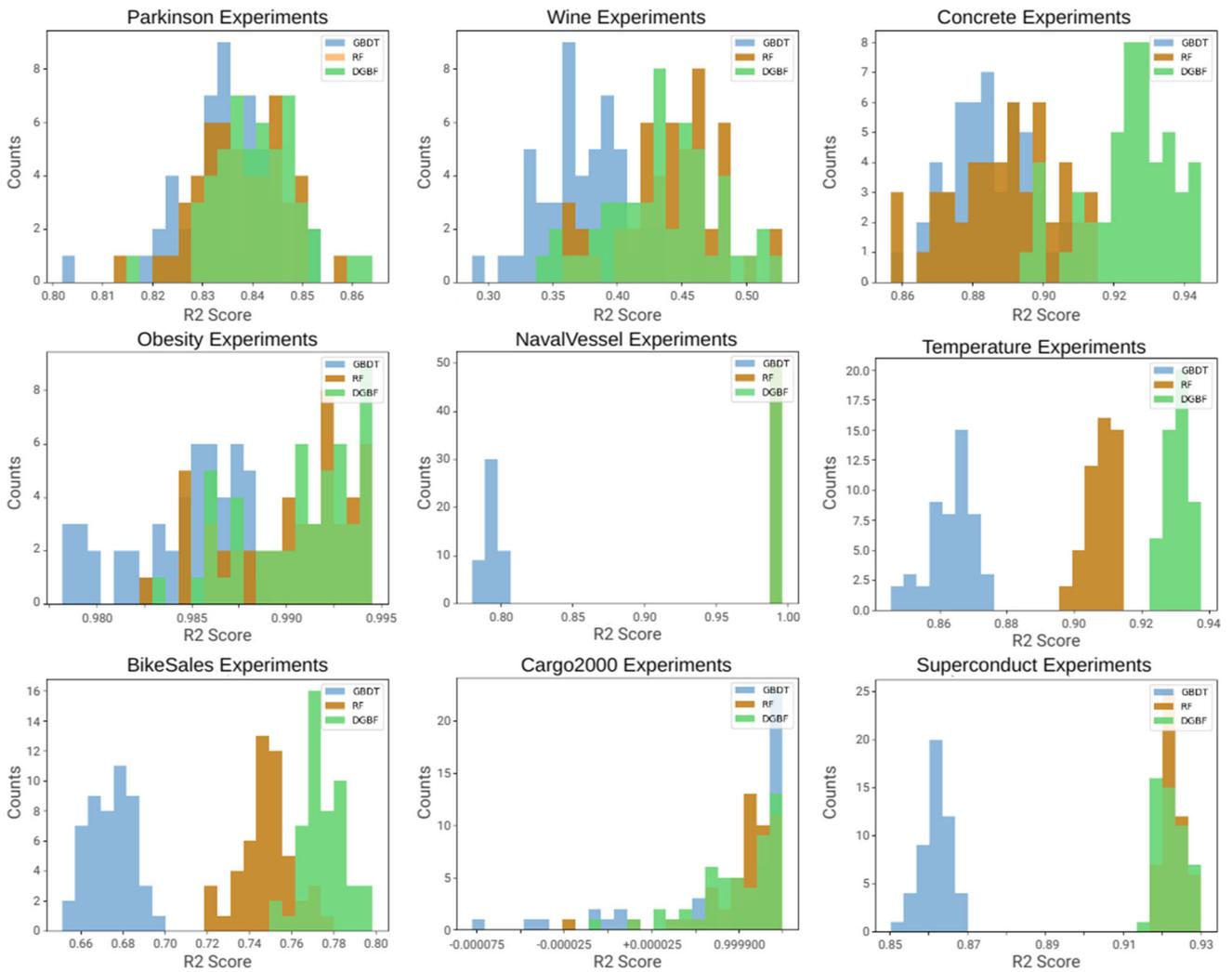

**Fig. 4** In this image it can be seen the distribution of R2 score of both models, RandomForest (brown), GradientBoosting (blue) and DGBF (green), for the 50 simulations of each dataset

**Fig. 5** In this image we meassure the execution time of each algorithm against the number of samples of the training set. Noting that the three algorithms are tree ensembles, the complexity is linearly related with the decision tree complexity, we plot the execution time againts the tree complexity ($n \cdot log(n)$)

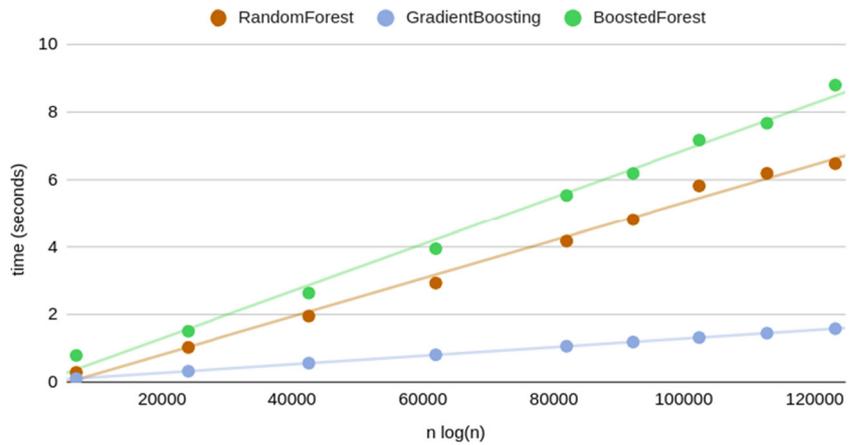

# Appendix A

# Appendix B background in CART, RandomForest and GradientBoosting bias

## B.1 CART bias

**Bias type 1** During the learning process, node regions are computed iteratively until an end condition is reached. Because of its exhaustive nature, the predictions tend to be biased producing high variance predictions overfitting the dataset $\{x_i\}$. This bias can be defined as

$$Bias_1(x) = E[Y \mid X = x] - E[y_i \mid x_i \in R_j] \mid_{x \in R_j} . \quad (16)$$

**Bias type 2** During the learning process, the splitting thresholds to produce the nodes are computed using only the middle point between two contiguous points from the dataset. No other value can be a threshold candidate. This produces a bias in the prediction due to two reasons. First because of lack of learning capacity in the underpopulated areas and second producing a high variance in the overpopulated areas. This can be visually understood in Fig. 6.

## B.2 Ensemble algorithms

Ensemble algorithms are able to solve the biased prediction of CARTs by combining the predictions of many models trained separately. There are different ensemble algorithms but the main two are GradientBoosting and RandomForest.

**RandomForest-Bagging** One way of combining the prediction of many trees to reduce bias is to aggregate the results by the mean. Given a dataset, the learning process of a tree is deterministic, so to make multiple trees produce different predictions, each of them is trained in a different bootstrap subsample of the dataset. This technique is called *bagging* [8].

$$F(x) = \frac{1}{n_{trees}} \sum_{j=0}^{n_{trees}} h_j(x), \quad (17)$$

where $h_j(x)$ is trained to minimize the loss function, $L(y, f(x))$, over a *bagging* subsample $\{x\}_j$

$$h_j(x) = \arg\min_h L(y_i, h(x_i))/x_i \in \{x\}_j. \quad (18)$$

This technique is based on the central limit theorem from statistics where we expect that the variance manifest of the CART biases can be reduced by averaging over enough tree predictions. All the trees learn in parallel, and this kind of learning is called "horizontal learning".

**GradientBoosting-Boosting** In contrast, in GradientBoosting each tree does not try to learn over a different dataset sample to minimize the loss, but it follows a "stage-wise" optimization approach, where each tree is added to the ensemble to reduce the global loss from the previous trees. The final prediction is the sum of all the trees in the ensemble. This process is called *boosting* [9]

$$F(x) = \sum_{m=0}^{M} h_m(x), \quad (19)$$

where $M$ is the number of trees of the ensemble. To reduce the loss function from the previous trees, $L(y, F_{t-1}(x))$, each of those trees is fitted with the gradient of the loss function

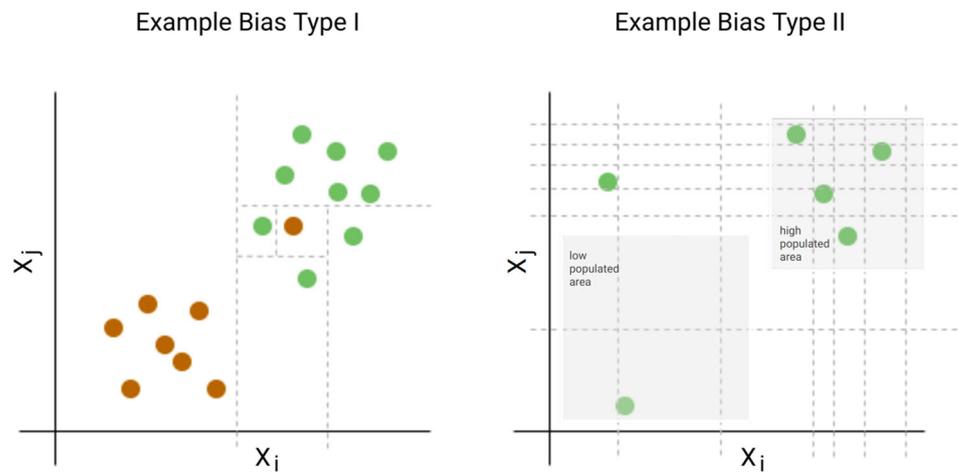

**Fig. 6** In the left there is an example of Bias type 1, where an outlier is clearly overfitted because of the exhaustive learning process. On the left, it can be seen the difference in the distribution of the possible threshold candidates in overpopulated and underpopulated areas

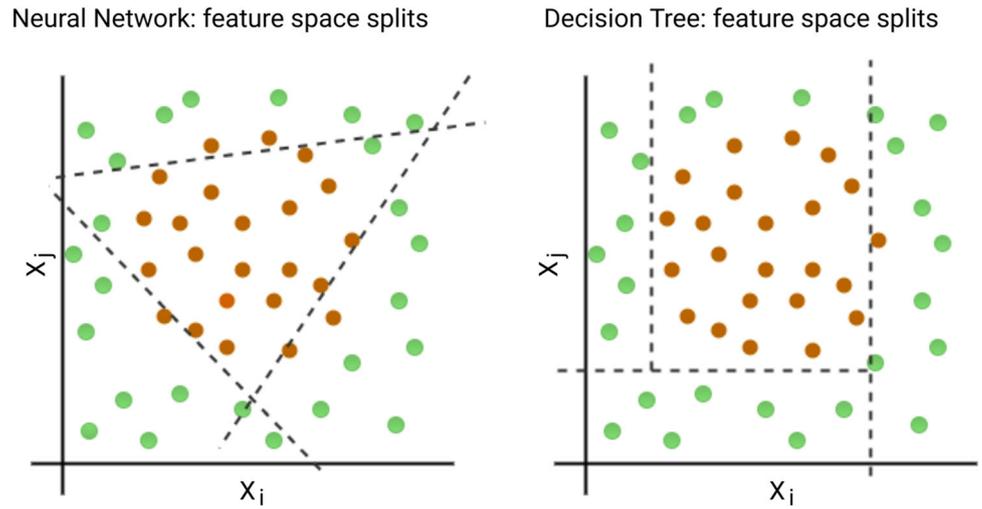

**Fig. 7** Given the same sample distribution of points in a feature space, the chart in the left shows the decision boundaries generated by Neural Networks algorithm with three neurons. The chart in the right shows the decision boundaries generated by a decision tree algorithm with three decisions

from the previous predictors:

$$h_m(x) = \rho_m g_m(x), \quad (20)$$

$$g_m(x) = E_y \left[ \frac{\partial L(y, F(x))}{\partial F(x)} \right]_{F=F_{m-1}},$$

$$\rho_m = \arg\min_{\rho'} E_{y,x} \left[ L(y, F_{m-1}(x) - \rho' g_m(x)) \right] \quad (21)$$

In the previous formula, the value of $E_y[.]$ can only be computed knowing the density function $P(y \mid x)$. In real-world problems, we never have the density function, consequently, the *boosting* is approximated using finite data by assuming regularity in the $P(y \mid x)$ distribution. To do so, each tree is trained with pseudo-responses that are computed (in the case of the $RMSE$ loss) as the difference between the target and the predictions of a current ensemble of trees (residual errors).

$$h_m(x) \simeq \arg\min_{h'} E \left[ L(y - F_{m-1}(x), h'(x)) \right], \quad (22)$$

While CART tries to reduce the loss during the training process by splitting leaf nodes into children nodes, the *boosting* algorithm tries to reduce the loss from each tree by adding another tree. However, the *boosting* algorithm generalizes better than the optimization from trees because the trees in each state, optimize the loss function using the entire dataset and not only the subsample from the previous node

### B.3 Bagging and boosting bias

Both, *bagging* and *boosting*, are different ensemble techniques that rely on different mathematical approaches. *Bagging* relies on generalization by reducing the variance by averaging the prediction from multiple predictors trained to predict the same response. Meanwhile, *boosting* relies on reducing the global training loss from the previous tree by training a new tree over the pseudo-responses. Even though these techniques are used separately, the two approaches are not incompatible.

**GradientBoosting Bias** In the finite data approximation, we suppose that training over the pseudo-responses is representative of the gradient. However, computing the pseudo-responses over the same training data from one tree to another produces biased pseudo-responses. It is demonstrated in [10] that the bias induced by the finite data approximation is:

$$Bias_{GB}(x) \equiv E[F_{t-1}(x')]_{x'=x} - E[F_{t-1}(x') \mid x' = x_k]_{x'=x}, \quad (23)$$

**RandomForest Bias** The reduction of bias from CART using the RandomForest ensemble is based on the reduction of the prediction variance by averaging the predictions from a high number of trees. However, averaging over all the trees does not ensure the convergence to a global minimum of the loss function everywhere (i.e. in all the areas of the feature space). This is due to the robustness of the decision trees to learn outlier regions, which is lost by averaging over the trees which have not been trained with the outliers.

$$Bias_{RF}(x) \equiv E \left[ h(x) \mid x \in R_j \right] - E \left[ h'(x) \mid x \in R'_j \right], \quad (24)$$

where $h(x)$ and $h'(x)$ are trees trained with different subsamples. If these subsamples have different statistics (for instance, because of a small outlier region), the ensemble is not flexible enough to learn this characteristic behavior of that region.

**Acknowledgements** I want to thank Sara San Luís Rodriguez for her selfless support and her perfectionism and also Bea Hernández Lorca because the help and questions she planted two years ago are the seeds of the trees from today.

**Data Availability** All the data and materials are available at https://doi.org/10.5281/zenodo.7236216.

## Declarations

**Competing interests** The authors declare that they have no conflicts of interest to this work. The people involved in the experiment have been informed and formally accepted.

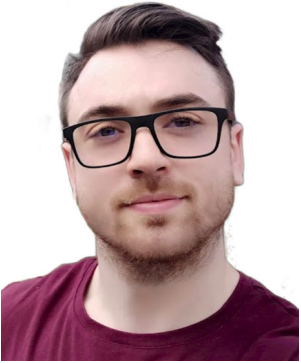

**Ángel Delgado-Panadero** received the B.S degree in Physics from the University of Salamanca, Spain. He has been working as an Artificial Intelligence Engineer in consultancy and software developing enterprises such as NTT and Indra. From 2020 he has also been working as a speaker, independent researcher and consultancy instructor such as EAE Business school.

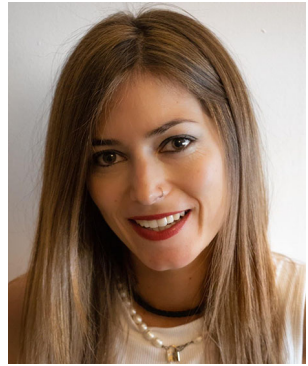

**María Teresa García-Ordás** Ph.D. was born in León, Spain, in 1988. She received her degree in Computer Science from the University of León in 2010, and her Ph.D. in Intelligent Systems in 2017. She was a recipient of a special mention award for the best doctoral thesis on digital transformation by Tecnalia. Since 2019, she works as Associate Professo at the University of León. Her research interests include computer vision and deep learning. She has published several articles in impact journals and patents. She has participated in many conferences all over the world.

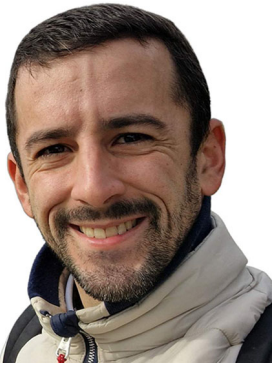

**José Alberto Benítez-Andrades** received the B.S. degree in Computer Engineering and PhD in Production Engineering and Computing from the University of León. He is an Associate Professor at the University of León. His research is related to the application of artificial intelligence techniques, knowledge engineering and social network analysis applied mainly to problems related to the field of health. He has more than 45 publications indexed in JCR, 30 communications in international conferences, is associate editor of the journal BMC Medical Informatics and Decision Making, has organized several international conferences since 2018 and is an evaluator of international projects for the government of Spain and Peru.